\documentclass[10pt,twocolumn,letterpaper]{article}

\usepackage{iccv}
\usepackage{times}
\usepackage{epsfig}
\usepackage{graphicx}
\usepackage{amsmath}
\usepackage{amssymb}

\usepackage{multirow}
\usepackage{makecell}
\usepackage{helvet}
\usepackage{courier}
\usepackage{graphicx}
\usepackage{bm}
\usepackage{color}
\usepackage{epstopdf}
\usepackage{caption}
\usepackage{subcaption}
\usepackage{enumitem}
\usepackage{calc}
\usepackage{multirow}
\usepackage{xspace}
\usepackage{booktabs}
\usepackage{mathrsfs}
\usepackage{array}
\usepackage{gensymb}
\usepackage{bbm}
\usepackage{dsfont}
\usepackage{bm}

\usepackage[breaklinks=true,bookmarks=false]{hyperref}

\iccvfinalcopy 

\def\iccvPaperID{****} 

\begin{document}

\title{Robust 2D/3D Vehicle Parsing in CVIS}


\author{Hui Miao\textsuperscript{1}, Feixiang Lu\textsuperscript{2}, Zongdai Liu\textsuperscript{1}, Liangjun Zhang\textsuperscript{2}, Dinesh Manocha\textsuperscript{2,3} and Bin Zhou\textsuperscript{1}\\\\
\textsuperscript{1}State Key Laboratory of Virtual Reality Technology and Systems, Beihang University\\
\textsuperscript{2}Robotics and Autonomous Driving Laboratory, Baidu Research \\
\textsuperscript{3}University of Maryland, College Park, MD 20742, USA
}

\maketitle

\begin{abstract}

   We present a novel approach to robustly detect and perceive vehicles in different camera views as part of a cooperative vehicle-infrastructure system (CVIS). Our formulation is designed for arbitrary camera views and makes no assumptions about intrinsic or extrinsic parameters. First, to deal with multi-view data scarcity, we propose a part-assisted novel view synthesis algorithm for data augmentation. We train a part-based texture inpainting network in a self-supervised manner. Then we render the textured model into the background image with the target 6-DoF pose. Second, to handle various camera parameters, we present a new method that produces dense mappings between image pixels and 3D points to perform robust 2D/3D vehicle parsing. Third, we build the first CVIS dataset for benchmarking, which annotates more than 1540 images (14017 instances) from real-world traffic scenarios. We combine these novel algorithms and datasets to develop a robust approach for 2D/3D vehicle parsing for CVIS.  In practice, our approach outperforms SOTA methods on 2D detection, instance segmentation, and 6-DoF pose estimation, by 4.5\%, 4.3\%, and 2.9\%, respectively.  More details and results are included in the supplement. To facilitate future research, we will release the source code and the dataset on GitHub.
   
   
   \end{abstract}
   
   \section{Introduction}
   \label{sec:intro}
   

   \begin{figure}
   \centering
   \includegraphics[width=0.98\linewidth]{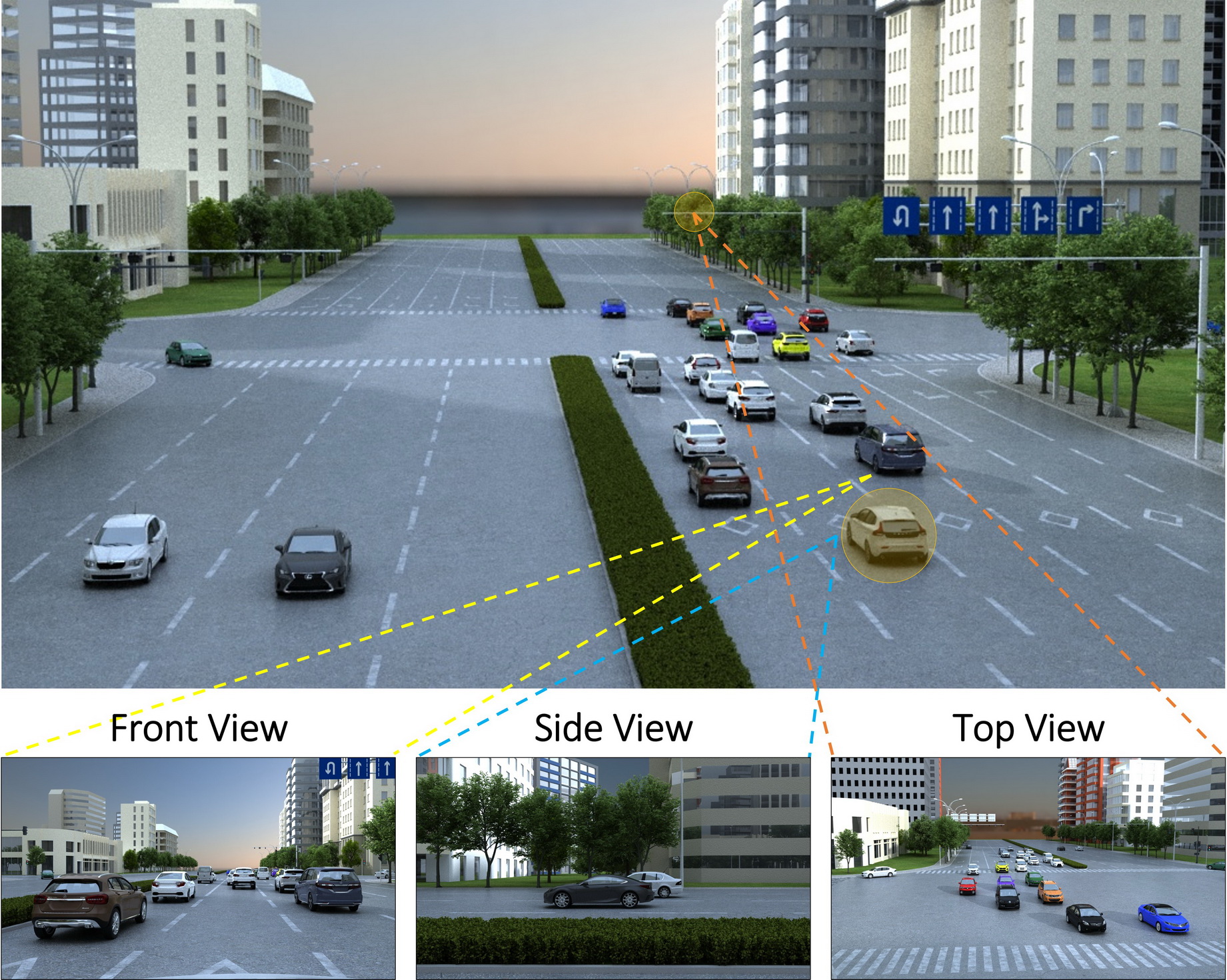}
   \caption{This is a perception diagram of a cooperative vehicle-infrastructure system (CVIS). Our goal is to re-use and augment the existing front-view AD datasets to robustly perform 2D/3D vehicle parsing in novel views (\eg, side view or top view).}
   \label{Fig:Teaser}
   \end{figure}
   
   Cooperative vehicle-infrastructure systems (CVIS) have become a key focus of research and technology in the field of autonomous driving (AD) \cite{siegel2017survey, reporterlinker2018cvis}. 
   In a CVIS, camera, radar, LiDAR and other sensors are mounted on  vehicles and street light poles at different locations (\eg, front, side, top \etc). The simultaneous perception of vehicles and road terminals can minimize blind zones and provide warning for out-of-sight collisions in advance. 
   An example is shown in Fig.~\ref{Fig:Teaser}, which is a typical traffic scenario. From the front-view of the autonomous vehicle, many of the other vehicles are partially or completely occluded, presenting perception challenges. From the side/top-view, we can clearly see many of these occluded vehicles views. 
   Although multiple views can get better perception results, in the end, performance relies on improvement \textit{\wrt} each of these single views.
   The ultimate perception problem in a CVIS, therefore, is \textit{how to effectively detect the vehicles in novel views}. Many widely used AD datasets (\eg, KITTI~\cite{geiger2013vision}, CityScapers~\cite{cordts2016cityscapes}, ApolloScape~\cite{huang2018apolloscape}, ApolloCar3D~\cite{song2019apollocar3d}) only provide labeled front-view data. If we directly use these labeled front-view datasets to train the deep neural networks (DNNs), the detection performance would degrade dramatically if we test these networks on the data from other views (\eg, the side view or the top-view).
   A common strategy is to capture new images for manually annotation, which is labor intensive, costly, and in-efficient.
   As a result, we need a novel set of view synthesis algorithms that can enable us to use current front-view AD datasets for CVIS.

   

   To address the data scarcity challenge, we propose augmenting existing AD datasets via novel view synthesis. In computer vision, view synthesis has been extensively studied. Researchers use 3D model rendering techniques (\eg, \cite{su2015render}, \cite{rematas2016novel}), image-based appearance flow approaches (\eg, \cite{zhou2016view}, \cite{zhu2018view}), or 
   generative adversarial network (GAN) (\eg, \cite{zhu2017unpaired}, \cite{sitzmann2019scene}) to synthesize novel-view images. Although they can achieve good results, these methods have several limitations in the context of CVIS. First, their synthesized results are mostly used for visualization, which are difficult to be learned by deep networks due to the domain gap. Second, it is difficult to obtain ground-truth annotations, especially 3D information (\eg, a 3D bounding box and a 6-DoF pose). Third, these approaches rely on multi-view or paired images as guidance for training, which can be obtained by 3D model rendering or a camera array in the laboratory setting. However, it is difficult to obtain such data in real AD scenarios due to occlusions and fast motions of vehicles.
   
   Some of these  synthesis problems can also be addressed using techniques for 3D parsing from single images. 3D parsing from a single image is important for AD, but challenging 
   because pinhole cameras cannot obtain absolute 3D positions due to projective mapping. Many state-of-the-art (SOTA) methods (\eg, \cite{ma2019accurate}, \cite{cai2020monocular}, \cite{ding2020learning}) require training a depth estimation network using the ground-truth depth map or stereo pairs, with fixed intrinsic and extrinsic parameters for the cameras. They can get good results on the training dataset, but their generalizability is limited because 1) it is difficult to obtain the depth map in the real world, especially the depth of background and 2) cameras have distinct intrinsic and extrinsic parameters. Therefore, it is difficult for the trained model in camera $\mathcal{A}$ to estimate the 6-DoF vehicle pose in camera $\mathcal{B}$.
   
   \subsection{Main Contributions}
   In this paper, we address the problem of re-using the existing front-view AD datasets to achieve novel-view 2D/3D vehicles parsing for CVIS. In order to re-use such datasets, we address two main challenges: 1) automatic view synthesis for data augmentation and ground-truth 2D/3D labels generation in novel views and 2) 3D parsing from a single image in novel views.
   We present a novel approach to detect and parse novel-view vehicles from a single image. This includes a new method for data augmentation of front-views in AD datasets and a novel approach for 2D/3D vehicle parsing. The key innovation of our data augmentation approach is the use of part-based texture inpainting for novel view synthesis. Specifically, we use the existing vehicle-based datasets that provide 3D vehicle templates and associated 6-DOF datasets. We first project the image pixels to the texture map of the template. Because there could be some holes or blank regions due to the projective mapping, we then train a part-based texture inpainting network to fill the missing data. After obtaining the complete texture map, we render the textured 3D templates with arbitrary 6-DoF poses into novel views. 
   
   Based on this synthesized data generated from vehicle templates, we present a new approach for 2D/3D parsing, that is robust for arbitrary camera parameters. Instead of directly learning depth from images, the key idea of our approach is learning dense mappings between the image pixels and the canonical 3D vehicle template. Specifically, we design a multi-task network that outputs results of 2D detection, instance segmentation, a 3D bounding box (\ie width, height, and length), and canonical 3D points. These canonical 3D points are one-to-one mapped to the image pixels; thus, we can compute the 6-DoF pose by solving the RANSAC-PnP problem using the input intrinsic and extrinsic parameters of the camera. 
   
   Finally, to benchmark our synthesized data and 2D/3D parsing approach, we have constructed, to the best of our knowledge, the first CVIS-oriented dataset with vehicles in different camera views, containing 1540 labeled images and 14017 2D vehicle instances. For each vehicle instance in our dataset, we annotate its 2D bounding box, instance-level segmentation, 3D bounding box, and 6-DoF pose. 

   In summary, our contributions include:
   

   

   1) We present a novel and compact data augmentation pipeline for vehicle parsing in CVIS. This includes synthesized data generation, data learning for 2D/3D parsing, and real-world dataset  construction for benchmarking.

   2) A 3D-assisted image augmentation approach is proposed to handle novel views. The key innovation is a part-based texture inpainting network with a self-supervised approach, which can automatically synthesize novel-view images with ground-truth annotations.

   3) For 2D/3D vehicle parsing, we learn dense correspondences between the image pixels and the 3D points in a 3D vehicle template, which is robust for 6-DoF pose estimation under arbitrary camera parameters.

   4) We build a dataset containing 1540 real images (14017 instances) with 2D/3D vehicle annotation in many different views. We compare with other SOTA methods and highlight the accuracy improvements.

   
   \section{Related Work}

   \subsection{Data Augmentation}
   The ``fuel'' of deep networks is labeled datasets.  The most common strategy for generating enough data to train a model is manually crowd sourcing real images to annotate, which is labor intensive~\cite{geirhos2018imagenet}. To decrease overhead and improve efficiency, researchers have recently focused on data augmentation techniques. Existing approaches can be categorized into four classes: 1) \textit{3D model-based approaches}; 2) \textit{image editing approaches}; 3) \textit{appearance flow-based approaches}, and 4) \textit{GAN-based approaches}.

   With the publication of large-scale 3D model datasets \cite{shapenet2015, dai2017scannet, mo2019partnet}, we can render these 3D models to generate images with ground-truth annotations such as view point \cite{su2015render}, segmentation \cite{rematas2016novel}, and depth map \cite{shotton2011real}.  Based on the rendering data, explicit methods \cite{kholgade20143d, chen20133, karsch2011rendering, zheng2012interactive, rematas2016novel, xin2018autosweep, kar2019meta} and implicit methods  \cite{yang2015weakly, varol2017learning, xu2019deep, zakharov2020autolabeling} are developed to synthesize the novel-view images. However, rendering is a time-consuming process that requires a lot of human interactions such as pre-building 3D scenes \cite{kaneva2011evaluation, engelmann2017exploring}. In addition, it is not easy to reduce the domain gap between the rendering data (CG style) and the real captured data (photo), though there are some domain adaptation methods (\eg, \cite{tobin2017domain}).

   Another approach for data augmentation is image editing \cite{liu20203d}. Dwibedi~\etal~\cite{dwibedi2017cut} cut objects from images and then paste them to other backgrounds to synthesize photo-realistic training data. However, these ``cut-paste'' data cannot rotate or translate in the 3D space. Moreover, this method is limited to handling occlusion problems.


   Appearance flow-based approaches \cite{zhou2016view, park2017transformation, zhu2018view} use the dense pixel-to-pixel correspondences to generate the novel-view images directly, which can be regarded as optical flow prediction. However, these methods are supervised on the target images, requiring controlled environments (\eg, camera pose, scene geometry). For AD, it is difficult to obtain such dense correspondences from street-view images.

   Recently, GAN-based approaches \cite{zhu2017unpaired, wang2018high, balakrishnan2018synthesizing, yao20183d, xu2019view, lv2020pose} have been proposed to generate photo-realistic images using a generator-discriminator architecture. However, GAN-based methods have some limitations. First, GAN results are not controllable or robust. Second, as with the majority of deep networks, their results largely depend on the training data. More importantly, it is difficult to generate ground-truth 3D annotations (\eg, 6-DoF pose) for training.

   \subsection{Image-based Vehicle Parsing in AD}
   
   For AD, it is important to parse the vehicles into 2D and 3D levels. Prior techniques can be divided into three categories: 1) \textit{2D detection} (\eg, SSD513~\cite{fu2017dssd}, YOLOv3~\cite{redmon2018yolov3}, Faster-RCNN~\cite{ren2015faster}); 2) \textit{instance-level segmentation} (\eg, Mask-RCNN~\cite{he2017mask}); and 3) \textit{6-DoF pose estimation or 3D detection} (\eg, AM3D~\cite{ma2019accurate}, DPOD~\cite{zakharov2019dpod}, PV-Net~\cite{peng2019pvnet}, D4LCN~\cite{ding2020learning}).
   These prior works can get good results in the current perspective, but their performance degrades in other novel views. In other words, it is difficult for their trained models in camera $\mathcal{A}$ to parse vehicles in camera $\mathcal{B}$.

   \subsection{Vehicle Re-ID}
   
    The work most related to ours is vehicle re-identification (Re-ID) (\eg, \cite{yang2015large}, \cite{liu2016deep}, \cite{tang2019pamtri}), which aims to identify all the images of vehicles with the same vehicle ID in different camera views. However,  the Re-ID images are cropped for classification/retrieval/categorization and do not have 2D instance or 6-DoF pose annotations. Thus existing vehicle Re-ID methods are limited to 2D/3D vehicle parsing such as instance-level segmentation and 6-DoF pose estimation, though Re-ID is an essential module in the context of CVIS.

   \subsection{Datasets for CVIS}
   
   Recently, several datasets have been constructed and released for 2D/3D perception in autonomous driving (\eg, KITTI~\cite{geiger2013vision}, CityScapes~\cite{cordts2016cityscapes}, Mapillary~\cite{neuhold2017mapillary}, ApolloScape~\cite{huang2018apolloscape}, and ApolloCar3D~\cite{song2019apollocar3d}). However, these data are captured from the front view, thus the trained model is limited to the test data on the side view or the top view. For CVIS, we focus more on 2D/3D vehicle perception from multiple camera views. The AI-City-Challenge \cite{Naphade21AIC21} provides city-scale multi-camera images/videos for vehicle Re-ID \cite{tang2019pamtri}, tracking \cite{Tang19CityFlow}, and retrieval \cite{Feng21CityFlowNL}, which are important to make transportation systems ``smart'' and efficient. However, these data and tasks focus more on 2D vehicle parsing \cite{Yao20VehicleX}, while our goal is 2D/3D parsing, particularly 6-DoF pose. We believe 3D information is critical to CVIS, enabling safer road and reducing traffic jams.


   

   
   \section{Method}
   
\begin{figure}[!htbp]
   \centering
   \includegraphics[width=0.98\linewidth]{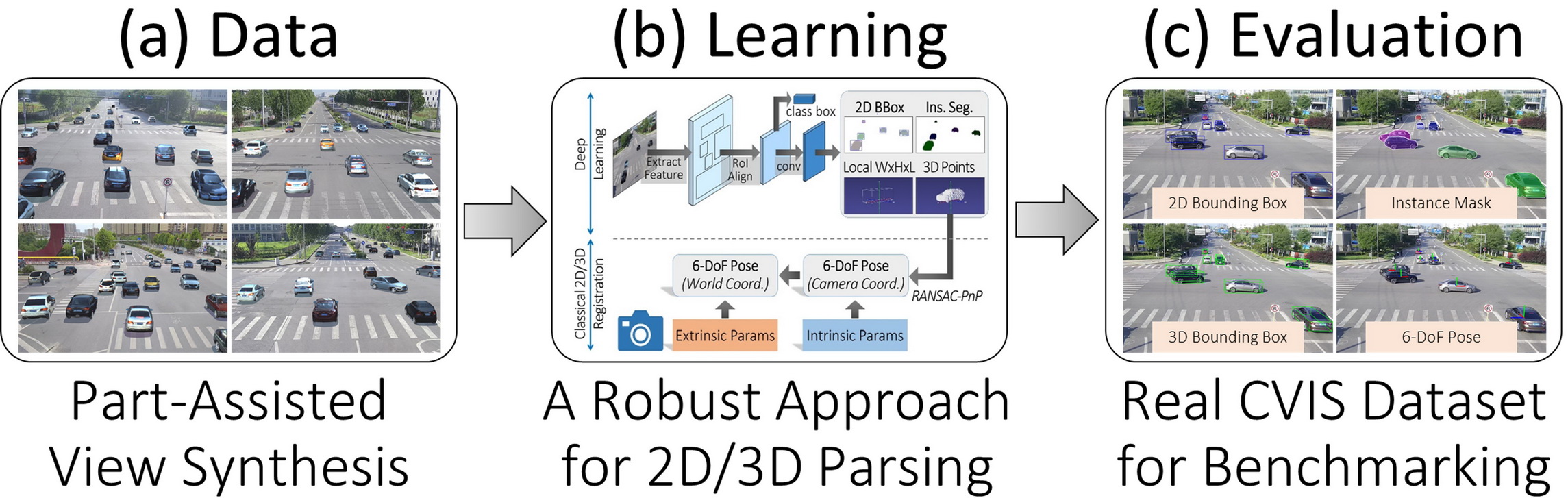}
   \caption{Three main components of our approach. }
   \label{fig:approach}
   \end{figure}

   In this section, we present a novel and compact data augmentation pipeline for 2D/3D vehicle parsing (shown in Fig.~\ref{fig:approach}), which includes three main components:

   1) \textit{Data} (Sec.~\ref{sec:view_syn}). We present a part-assisted data augmentation method  to synthesize images and generate ground-truth annotations for training network. The key innovation is part-based inpainting network (Sec.~\ref{sec:inpainting network}).

   2) \textit{Learning} (Sec.~\ref{sec:parsing}). Based on our synthesized data, we present a robust approach to perform 2D/3D parsing.

   3) \textit{Evaluation} (Sec.~\ref{sec:dataset}). We construct a real-world, novel dataset for benchmarking on 2D/3D tasks.

   \begin{figure*}
   \centering
   \includegraphics[width=0.98\linewidth]{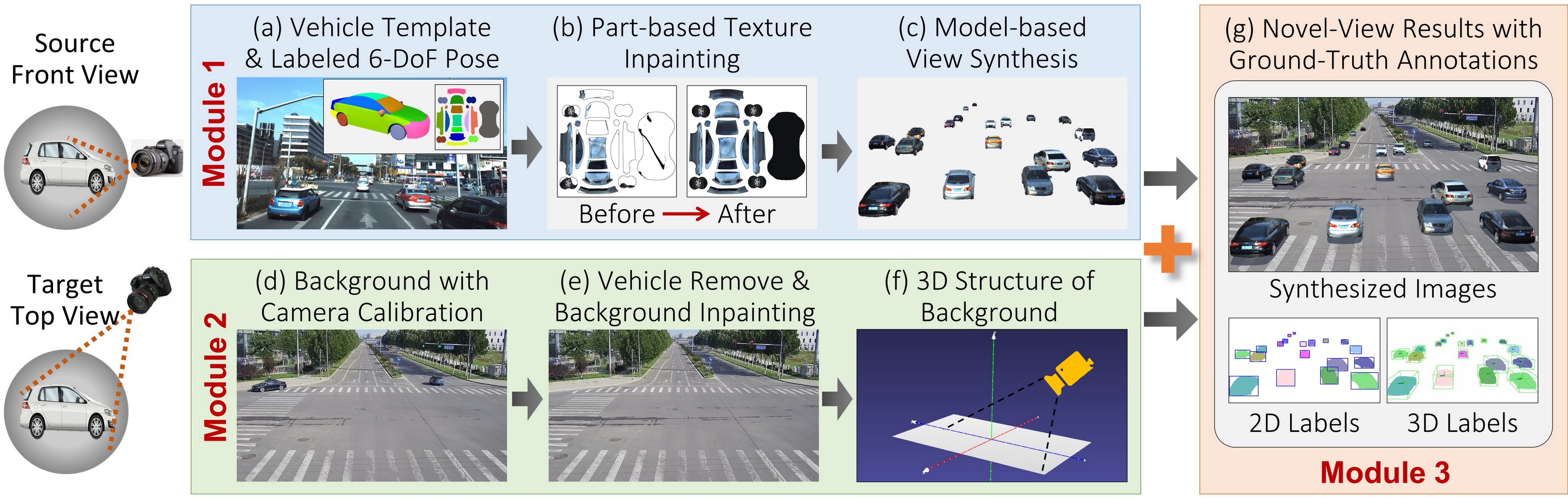}
   \caption{Our view synthesis approach includes three main modules: 1) \textit{Foreground}. (a) We use a part-level vehicle template representation to synthesize the foreground (\ie vehicles). Based on the labeled 6-DoF pose between the 2D instance and the 3D template, we can project the image pixels to the texture map with missing regions. (b)  A part-based texture inpainting network is proposed to synthesize a complete texture map. (c) Then we can render the textured vehicle template to generate novel-view images. 2) \textit{Background}. (d) We use the real traffic images with the pre-calibrated intrinsic and extrinsic parameters. (e) The moving objects are inpainted to generate a clean background. (f) We further compute the 3D structure of the background using the camera parameters. 3) \textit{Annotations}. (g) We render the textured vehicle template to the background to synthesize color images with ground-truth 2D/3D annotations.}
   \label{Fig:Overview}
   \end{figure*}
   
   \subsection{Part-Assisted View Synthesis}
   \label{sec:view_syn}

   To generate the novel-view data for training network, we propose a novel part-assisted view synthesis approach (Fig.~\ref{Fig:Overview}). The inputs are the 3D vehicle template and associated 6-DoF pose datasets (\textit{e.g.}, \cite{song2019apollocar3d}, \cite{lu2020permo}); and the background images with camera calibration. The output is the synthesized novel-view images with ground-truth 2D/3D annotations. Our approach can be divided into three modules: \textit{1) foreground}, \textit{2) background}, and \textit{3) annotations}.

   For the foreground, we focus on vehicle synthesis in this paper. The input is the labeled 2D-3D vehicle dataset, which annotates the 6-DoF pose and the  real-size 3D shape (deformed from parametric geometric vehicle representation) for each 2D vehicle instance. As shown in the top-right of Fig.~\ref{Fig:Overview} (a), the parametric geometric vehicle representation has two features. First, it is a \textit{geometric model} with an unfolded texture map and multiple uniform vehicle parts (\eg, door, body, trunk). Second, it is also a \textit{parametric model} with the PCA representation, where any new vehicle model can be represented as a linear combination of several principal components with coefficients. Based on the labeled 6-DoF pose and the camera parameters, we can project the 3D template to the image plane. Thus the image pixels can be mapped to the texture map of the 3D template, which inevitably yields a lot of missing regions due to projective mapping (Fig.~\ref{Fig:Overview} (b)). Then we design a part-based texture inpainting network to fill these missing regions (details in Sec.~\ref{sec:inpainting network}). After that, we can obtain complete texture maps for 3D model rendering with different intrinsic and extrinsic parameters of cameras under various illuminations (Fig.~\ref{Fig:Overview} (c)). Note that our vehicle data augmentation approach is generic and can be used to handle pedestrians. Such objects can be constructed from parts and we can use part-based PCA templates (\eg, \textit{SMPL} \cite{loper2015smpl}). The pedestrians augmentation results are shown in the supplement.

   For the background, we capture a lot of images from multiple views, especially the side-view and the top-view, where the camera's intrinsic and extrinsic parameters are pre-calibrated (Fig.~\ref{Fig:Overview} (d)). Next, we remove the existing vehicles through the inpainting method (\ie \cite{zhang2020autoremover}) to generate a clean background (Fig.~\ref{Fig:Overview} (e)). Then we estimate the 3D structure of the background (\ie the normal of the road plane) based on the camera's intrinsic and extrinsic parameters (Fig.~\ref{Fig:Overview} (f)).

   Based on  the ``clean'' background image and the associated 3D structure (module 2), we randomly put the 3D textured vehicles (module 1) with collision avoidance through intersection detection of 3D bounding boxes. Then we render the posed vehicle model to the background images. To enhance the fidelity of the synthesized results, we further generate the vehicle shadows according to the environment illumination and the road plane (Fig.~\ref{Fig:Overview} (g)). In contrast to existing view synthesis methods for visualization, our approach can  generate the ground-truth 2D/3D annotations, including 2D bounding boxes, instance-level segmentation, 3D bounding boxes, and 6-DoF poses. Benefiting from the 3D vehicle template, we can further generate the dense mapping data between the image pixels and the 3D template vertices according to the 6-DoF pose. We take these dense mapping data as a bridge to perform vehicle parsing from the 2D space to the 3D space (details in Sec.~\ref{sec:parsing}).
   
   

   \subsubsection{Part-Based Texture Inpainting Network}
   \label{sec:inpainting network}
   
   From the 3D vehicle template, we propose a novel part-level texture inpainting network which has two advantages. First, our network does not need paired multi-view images as guidance and the input texture maps are directly projected from real-world images. Second, we use a graph-based module to learn the texture inpainting features, and this module not only encodes the individual parts but also maintains the consistency among different parts.

   
   \begin{figure}
      \centering
      \includegraphics[width=0.98\linewidth]{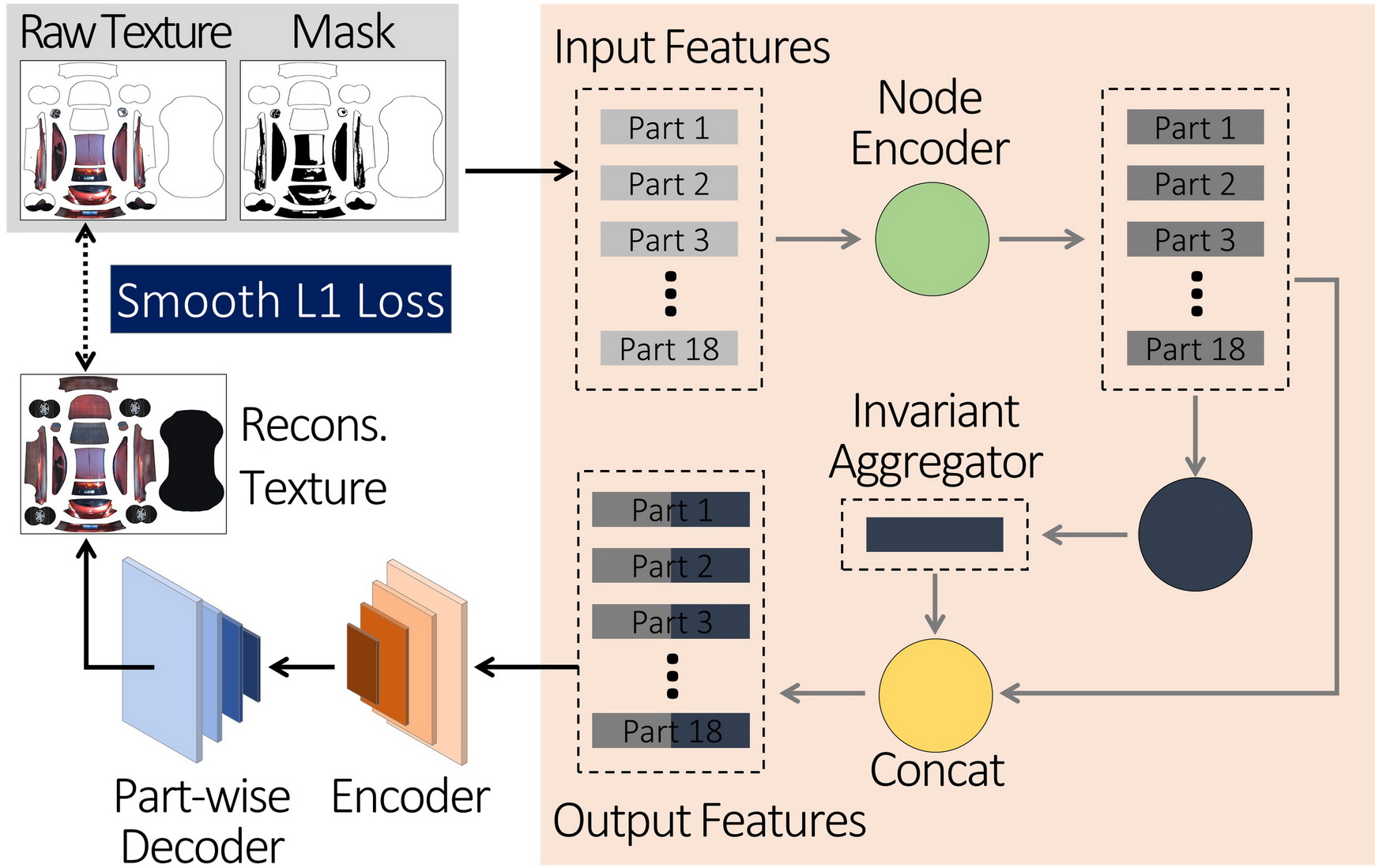}
      \caption{Our part-based texture inpainting network. We design an encoder-decoder network with the  graph-based inpainting module using the smooth L1 loss.}
      \label{Fig:InpaintingNetwork}
   \end{figure}

   Fig.~\ref{Fig:InpaintingNetwork} shows our inpainting network, which adopts an encoder-decoder architecture. Specifically, in the encoder stage, our network aggregates local features from the individual parts and then propagates to other parts for completion. As highlighted in Sec.~\ref{sec:view_syn}, the texture map $P$ has multiple parts (\ie 18) $\left \{ p_1, p_2, ..., p_{18} \right \}$. Each part can be regarded as a node in the graph. Then we define a single layer of graph propagation operation as 
   \begin{equation}
   p^{\left ( l+1 \right )}_i=\varphi_{rel} \left \{ g_{enc}\left ( p^ {\left ( l \right )}_i  \right ), \varphi_{agg}\left [  g_{enc}\left ( p_j^{\left ( l \right )} \right )  \right ] \right  \},
   \end{equation}
   where $p^{(l)}_i$ indicates the node feature in the $l$-th layer of the graph network and $p^{(0)}_i$ is the input patch of the image. The function $g_{enc}\left ( \cdot  \right )$ encodes the individual node/part features, $\varphi _{agg}\left [ \cdot  \right ]$ aggregates the features of all nodes (\ie 1$\sim$18), and $\varphi _{rel}\left \{ \cdot  \right \}$ is the relational operator between node/part $p_i$ and the aggregation of all nodes. In the decoder stage, we use a part-level decoder to recover each part from the last node/part feature $p_i^{last}$, which is defined as
   \begin{equation}
   \tilde{p}_i= g_{dec}^i\left ( p_i^{last} \right ).
   \end{equation}
   
   In our implementation, $g _{enc}\left ( \cdot  \right )$ is a convolutional layer followed by batch normalization and then ReLU non-linearity. $\varphi _{agg}\left [ \cdot  \right ]$ is the part axis maxpooling. $\varphi _{rel}\left \{ \cdot  \right \}$ concatenates the aggregated feature to each part. $g _{dec}\left ( \cdot  \right )$ is a sequential deconvolutional layer to recover the $i$-th part image. We use a four-level part-wise graph network to encode the relations among all parts. In the training stage, we randomly mask the existing part patches to train the network to recover them in a self-supervised manner. We compute the smooth-L1 loss between $p_i$ and $\tilde{p}_i$ for all existing parts.

    \begin{figure}
   \centering
   \includegraphics[width=0.98\linewidth]{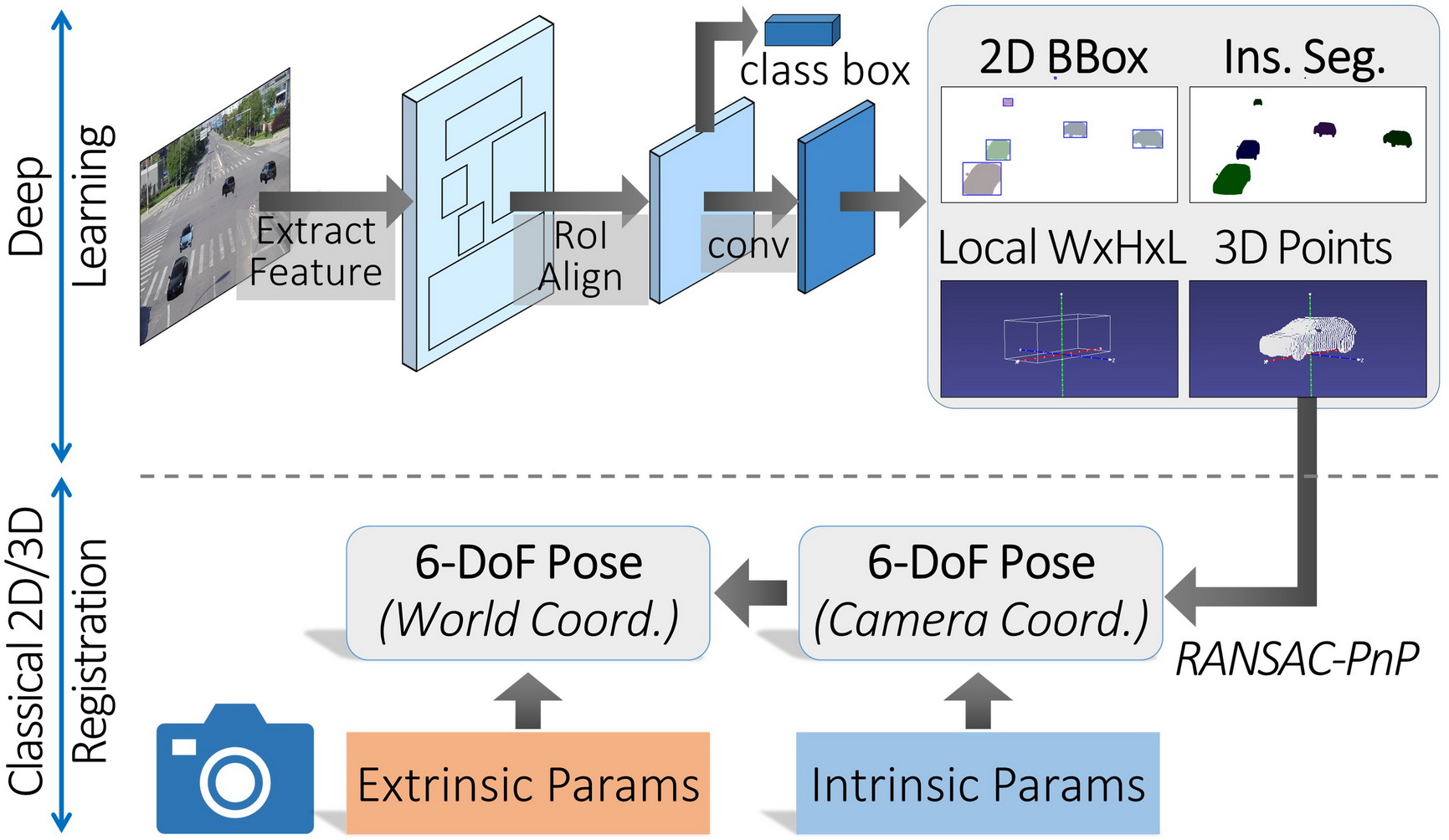}
   \caption{ Robust 2D/3D vehicle parsing in different camera views using our synthesized data. We train a multi-task deep network and then utilize classical 2D/3D registration techniques to perform the 2D/3D vehicle parsing. }
   \label{Fig:ParsingNetwork}
   \end{figure}


   
   \subsection{A Robust Approach for 2D/3D Parsing}
   \label{sec:parsing}
   
   Estimating 3D information from a single image is a very challenging task due to projective mapping. Thanks to the synthesized data with ground-truth 6-DoF pose and real-size 3D shape (Sec.~\ref{sec:view_syn}), we can perform 2D/3D vehicle parsing in different camera views. Specifically, our approach combines the strengths of deep learning and the classical 2D/3D registration techniques, which are robust for arbitrary camera intrinsic and extrinsic parameters. 
   The key idea is producing dense 2D/3D mappings between the image pixels and the canonical 3D template vertices. More specifically, we propose a multi-task network to learn the synthesized images with ground-truth annotations. As shown in Fig.~\ref{Fig:ParsingNetwork}, our network can output the parsing results of 2D detection, instance-level segmentation, a 3D bounding box (width, height, length), and dense 3D points.

   In contrast to existing ``key-points''-based pose estimation approaches (\eg, ApolloCar3D~\cite{song2019apollocar3d}), the key to our network is directly regressing each pixel’s 3D points of the vehicle template in the canonical space. Specifically, for each pixel $c_i = (u_i, v_i)^\top$ in the region of interest ($RoI$), we estimate its 3D point $v_i = (x_i, y_i, z_i)^\top$ in the canonical space. This process can be formulated as
   \begin{equation}
   \tilde{V} = Regressor(C),
   \end{equation}
   where $C = \left \{ c_1,c_2,c_3,...,c_n  \right \}$ and $\tilde{V} = \left \{ \tilde{v}_1,\tilde{v}_2,\tilde{v}_3,...,\tilde{v}_n  \right \}$. Based on the predicted 3D points, we formulate the 6-DoF pose estimation as a typical 2D/3D registration problem, which can be defined as 
   \begin{equation}
   Pose=\kappa \left ( \tilde{V}, C, K_{int}, K_{ext} \right ),
   \end{equation}
   where $K_{int}$ and $K_{ext}$ indicate the camera's intrinsic and extrinsic parameters, respectively.  $\kappa$ is the solver of the 2D/3D registration such as Efficient-PnP and RANSAC-PnP.

   In our implementation, we first use the 3D vehicle template to generate dense ground-truth mappings between $C$ and $V$ according to the 6-DoF poses. Then we integrate our canonical point regression module into Mask-RCNN~\cite{he2017mask}  as a new branch. In addition, we add another branch to estimate the dimension $[w, h, l]$ of the vehicle. As a result, our new multi-task network can detect/segment a vehicle and regress its 3D points and 3D dimension. The loss function of 3D points is defined as
   \begin{equation}
   L_{local3d}=SmoothL1\left ( \tilde{V}, V \right ),
   \end{equation}
   and the loss function of 3D dimension is defined as
   \begin{equation}
   L_{dimen}=SmoothL1\left ( \left [ \tilde{w}, \tilde{h}, \tilde{l} \right ],\left [ w,h,l \right ] \right ).
   \end{equation}

   
   \subsection{Dataset for Benchmarking}
   \label{sec:dataset}
   
   To the best of our knowledge, none of the existing datasets provide detailed 2D/3D annotation of vehicles from different views in real traffic scenarios. In this paper, we construct the first 2D/3D vehicle parsing dataset in CVIS for testing. Specifically, we capture source data from 20 street lights in three different cities, where 1 camera and 1 Lidar are mounted at each street light. The intrinsic and extrinsic parameters of camera and LiDAR are pre-calibrated. Then we manually annotate the 2D bounding box and instance mask for each vehicle in the image. The associated 3D bounding box and the 6-DoF pose are labeled on the 3D point clouds. In summary, our dataset contains 14017 annotated car instances from 1540 images in different views.

   The size of our dataset is comparable to other datasets (\textit{e.g.}, \textit{CityScapes} (1525 images), \textit{ApolloCar3D} (1041 images) for testing). We will release our dataset along with source code to facilitate future research. Note that our trained deep model is used directly for testing without any ``fine-tuning'' or ``mix-data-training'' strategies, while the training data consists of only synthesized images.


   \section{Experiments and Discussion}
   
\begin{figure}
   \centering
   \includegraphics[width=0.98\linewidth]{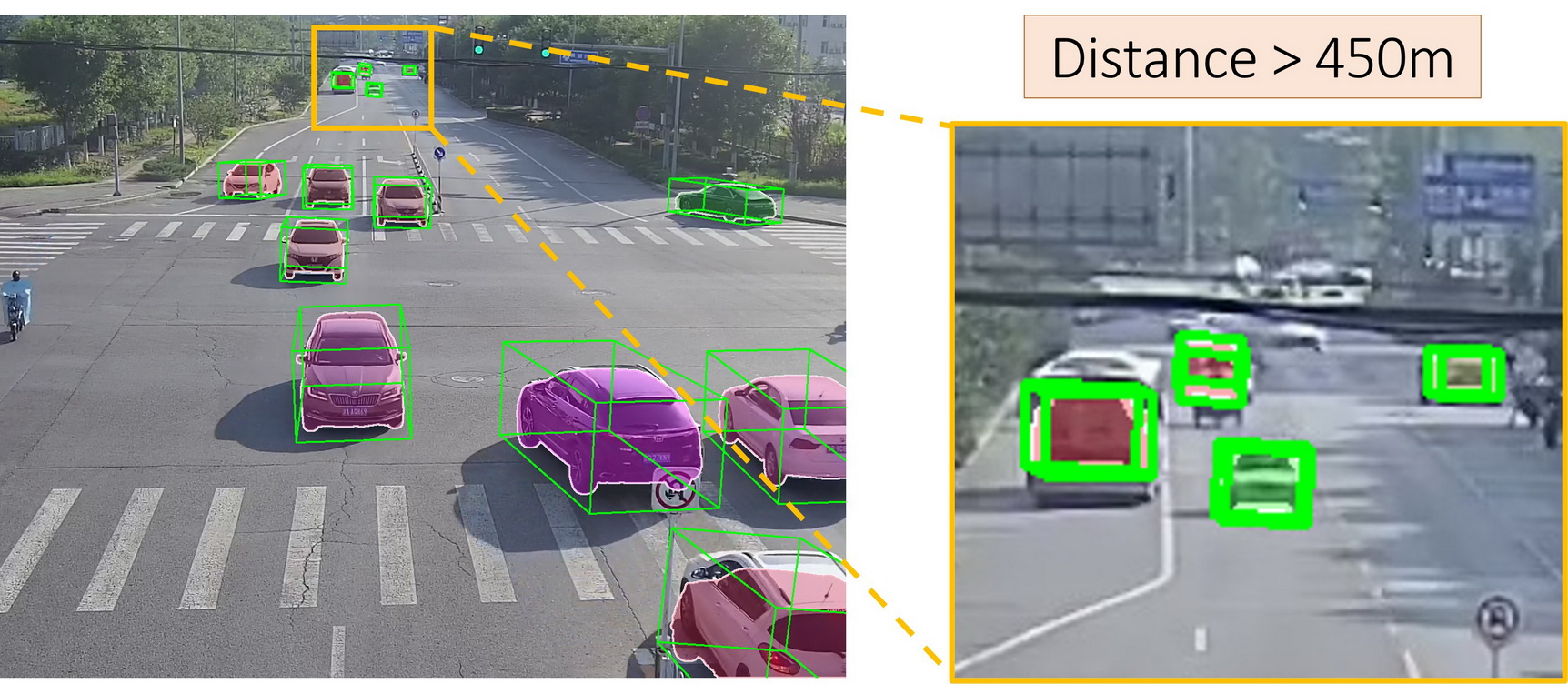}
   \caption{The 2D/3D parsing results using our approach, which is robust even for long range perception (\textgreater 450m).}
   \label{Fig:ParsingResults}
   \end{figure}

   Our approach can robustly output 2D/3D parsing results under arbitrary camera parameters. Owing to the high-quality synthesized data and the efficient parsing network, our approach performs well even for long range perception, where the vehicles have very small footprints (Fig.~\ref{Fig:ParsingResults}). \textit{More results (i.e. inpainted textures, synthesized images, 2D/3D paring results) are included in the supplement}. 
   
   \subsection{Implementation Details and Computation Time}
   
   We use the PerMO dataset~\cite{lu2020permo} and the ApolloCar3D dataset~\cite{song2019apollocar3d} to generate our synthesized data. Specifically, PerMO provided a part-level 3D deformable vehicle template with an unfolded texture map, and ApolloCar3D labeled the 6-DoF pose for each vehicle instance. The runtime for each synthesized vehicle is about 1.07 seconds. Specifically, it takes 0.07s for texture inpainting, 0.5s for background processing, and 0.5s for textured model rendering.
   
   
   The training time of our network depends on the data number. In general, training 20K images costs 16 hours (8 Nvidia P40 graphics cards). In the testing phase, we directly use the trained model to perform 2D and 3D parsing on our dataset. For an image with 10 vehicles, the average runtime is 0.23s. Specifically, it takes 0.2s for network prediction and 0.03s for 6-DoF pose estimation.  
   

   \subsection{Evaluation Metric}
   \label{sec:metric}
   
   We use $mAP$ as an evaluative criterion for 2D detection and instance-level segmentation because it is commonly used in many perception tasks \cite{Everingham10}.
   
   We follow the evaluation criteria proposed in ApolloCar3D~\cite{song2019apollocar3d} for 6-DoF pose evaluation, where \textit{``A3DP-Abs''} (\textit{``A3DP-Rel''}) means the absolute (relative) distance criterion. In addition, \textit{``c-l''} (\textit{``c-s''}) indicates results from a loose (strict) criterion. More details are introduced in \cite{song2019apollocar3d}.
   
   \subsection{Comparisons}
   
   \subsubsection{Comparison with Data Augmentation Methods}
   To demonstrate that our synthesized data can effectively improve the network performance, we compare it with other data, including 1) \textit{the existing dataset}, 2) \textit{rendering data}, and 3) \textit{GAN-based data}. Specifically, we use  \textit{ApolloCar3D}~\cite{song2019apollocar3d} as the existing dataset, which provides the vehicle annotations in the front view. We follow the approach \cite{liu20203d} to obtain the rendering data using the 3dsMax software. The GAN-based data is generated by the approach from \cite{lv2020pose}, which can synthesize novel-view vehicles according to the pre-defined poses. More details  in the supplement.
   
   To perform a fair comparison, we maintain the same number of training data to train the \textit{Faster-RCNN}~\cite{ren2015faster} network on the 2D detection task, the \textit{Mask-RCNN}~\cite{he2017mask} network on the instance-level segmentation task, and the \textit{DensePose}~\cite{alp2018densepose} network on the 6-DoF pose estimation task. As shown in Tab.~\ref{tab::comparison_data}, our synthesized data outperforms other data by a large margin. Specifically, the existing dataset only labels the front-view data, and thus the trained model is limited to other views. Rendering data has the natural domain gap with the real images, resulting in low performance. In addition, 3D rendering costs much more than 5x our data augmentation approach. Although the GAN-based data look more photo-realistic, their extracted features are different from the real images in AD, limiting the network training. More visualization results are in the supplement.
   

   \begin{table}
     \begin{center}
      \begin{tabular}{c|c|c|c}
      \toprule[1pt]
      \multirow{2}{*}{\textbf{Methods}} & \textbf{2D Det.} & \textbf{Ins. Seg.} & \textbf{6-DoF Pose}\\
      & (\textit{mAP}) & (\textit{mAP}) & (\textit{Abs-mean}) \\
      \hline
      \textit{Existing Dataset} & 48.7 & 48.1 & 22.4 \\
      
      \textit{Rendering Data} & 46.5 & 45.7 & 20.2 \\	
      
      \textit{GAN-based Data} & 39.1 & 37.4 & / \\
      
      \textbf{\textit{Our Data}} & \textbf{53.2} & \textbf{52.4} & \textbf{25.3} \\
      \bottomrule[1pt]
      \end{tabular}
      \vspace{-0.05in}
      \caption{ 2D/3D results with different data augmentation approaches. Note that these tasks are evaluated by SOTA methods (\ie \textit{Faster-RCNN} for 2D detection, \textit{Mask-RCNN} for instance segmentation, and \textit{DensePose} for 6-DoF pose).}
       \label{tab::comparison_data}
     \end{center}
   \end{table}

   \subsubsection{Comparison with Hole Filling Methods}
\begin{figure}
   \centering
   \includegraphics[width=0.98\linewidth]{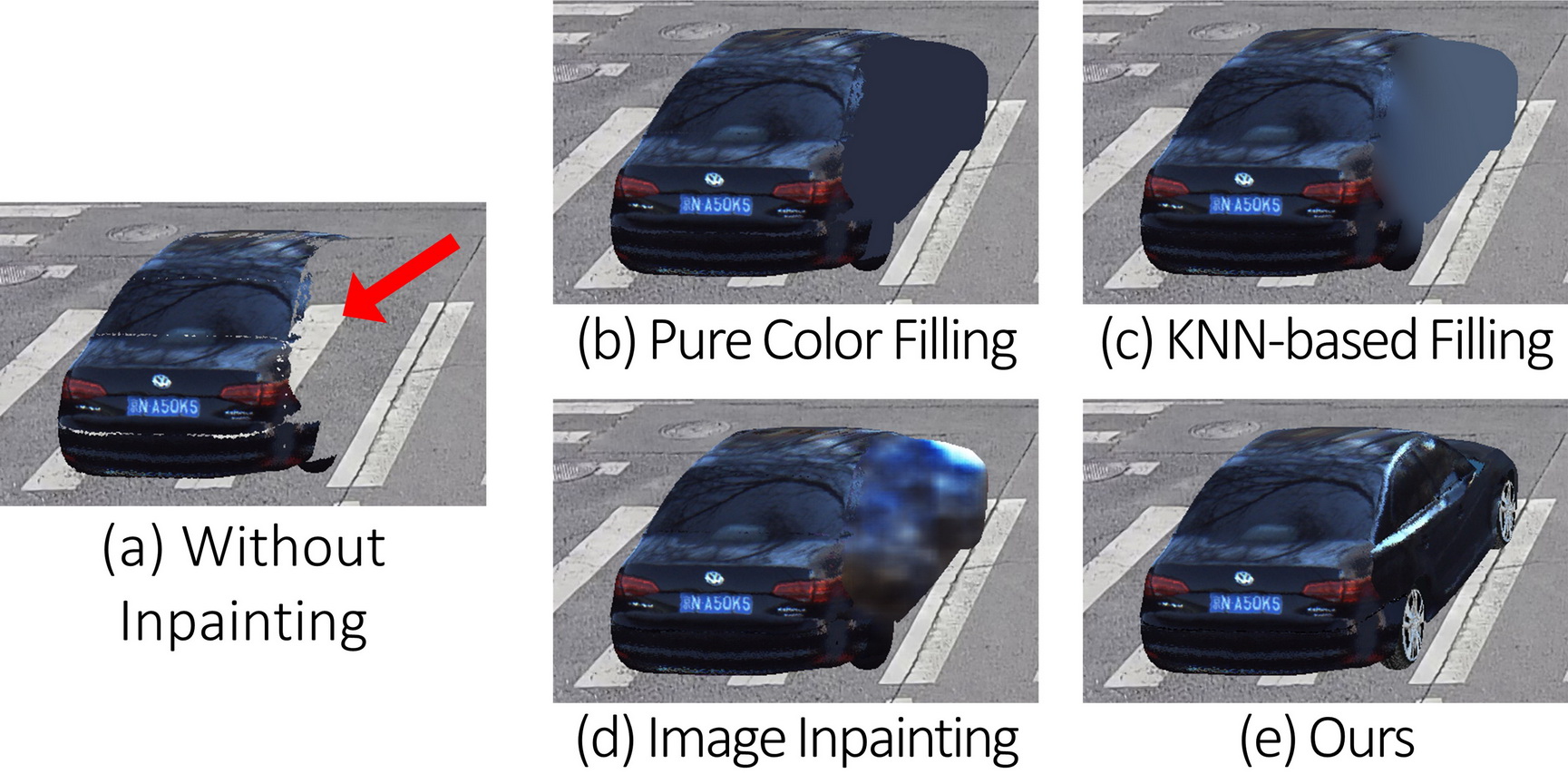}
   \caption{Visualization results of our part-based inpainting approach and other hole filling methods.}
   \label{Fig:InpaintingCompare}
   \end{figure}

   To justify the effectiveness of our part-based texture inpainting approach, the following hole filling methods are compared. 
   1)~\textit{Pure Color Filling.} We directly fill the missing regions using a pure color, which is a mean value of the existing region. 
   2)~\textit{KNN Filling.} We fill the missing regions using a linear blending algorithm where each pixel value is determined by the K-nearest neighbors (KNN). 
   3)~\textit{Image-based Inpainting.} We use the vehicle images and  randomly generate the missing regions to train an encoder-decoder network to fill the missing regions. 
   4)~\textit{Ours: Part-based Texture Inpainting.} We train a part-based inpainting network on the texture map.
   Here, we use our network (Sec.~\ref{sec:parsing}) to train these data. Tab.~\ref{tab:comparison_filling} shows that our approach outperforms the second best results on 2D detection and instance segmentation by 1.8\% and 2.5\%, respectively. The filling results of different approaches are shown in Fig.~\ref{Fig:InpaintingCompare}. Our approach can inpaint the missing regions while maintaining the color and structure consistency.

   \begin{table}
     \begin{center}
      \begin{tabular}{c|c|c}
      \toprule[1pt]
      \textbf{Methods}& \textbf{2D Det.} (\textit{mAP}) & \textbf{Ins. Seg.} (\textit{mAP})\\    
      \hline
      \textit{Pure Color} & 55.0 & 50.6 \\
      
      \textit{KNN Filling} & 57.5 & 51.1 \\	
      
      \textit{Image Inpainting} & 55.3 & 47.4 \\
      
      \textbf{\textit{Ours}} & \textbf{59.7} & \textbf{54.6} \\	
      \bottomrule[1pt]
      \end{tabular}
      \vspace{-0.05in}
      \caption{ We use our network (Sec.~\ref{sec:parsing}) to quantitatively compare our inpainting method with other inpainting methods on the tasks of 2D detection and instance segmentation. }
       \label{tab:comparison_filling}
     \end{center}
   \end{table}
   

   \subsubsection{Comparison with Pose Estimation Methods}
   Existing pose estimation methods can be categorized into three classes: 1) \textit{``Direct'' methods}, which directly regress the vehicle poses from the 2D images (\eg, 3D-RCNN~\cite{kundu20183d}); 2) \textit{``Key-points''-based methods}, which extract the pre-defined key-points and then solve a PnP problem to obtain the 6-DoF pose (\eg, DeepMANTA~\cite{chabot2017deep}, ApolloCar3D~\cite{song2019apollocar3d}); and 3) \textit{``Dense mapping''-based methods}, which regress the dense mapping between the 2D image and the UV map of the 3D model before solving the 6-DoF pose. (\eg, DensePose~\cite{alp2018densepose}). As reported by \cite{song2019apollocar3d}, ApolloCar3D advances 3D-RCNN and DeepMANTA by a big margin. Therefore, we retrain the ApolloCar3D network and the DensePose network using the same training data for fair comparison. 
   Tab.~\ref{tab:comparison_pose} shows the 6-DoF pose estimation results. Our approach outperforms the ApolloCar3D and DensePose on ``Abs-mean'' by 9.7\% and 7.7\%, respectively.

   \begin{table}
   \centering
   \addtolength{\tabcolsep}{-1.5pt}
   \begin{tabular}{l|lll|lll} 
   \toprule[1pt]
   \multirow{2}{*}{\textbf{Methods}} & \multicolumn{3}{c|}{\textbf{A3DP-Abs}} & \multicolumn{3}{c}{\textbf{A3DP-Rel}}  \\
                            & \textit{mean} & \textit{c-l} & \textit{c-s}              &  \textit{mean} &  \textit{c-l} &  \textit{c-s}              \\ 
   \hline
   
   \textit{ApolloCar3D} &$23.3$ & $31.7$ & $26.7$ & $22.5$ & $28.7$& $23.8$  \\
   
   \textit{DensePose} &$25.3$ & $35.6$ & $28.7$ & $25.1$ & $32.7$& $26.7$  \\
   \textbf{\textit{Our Method}} &  $\textbf{33.0}$ & $\textbf{47.5}$ & $\textbf{38.6}$ & $\textbf{27.8}$ & $\textbf{38.6}$ & $\textbf{29.7}$\\
   \bottomrule[1pt]
   \end{tabular}
   \vspace{-0.05in}
   \caption{6-DoF pose evaluation with different approaches. We introduce the evaluation metric in Sec.~\ref{sec:metric}. }
   \label{tab:comparison_pose}
   \end{table}
   
   \subsubsection{Comparison with 3D Detection Methods}
\begin{figure}
   \centering
   \includegraphics[width=0.98\linewidth]{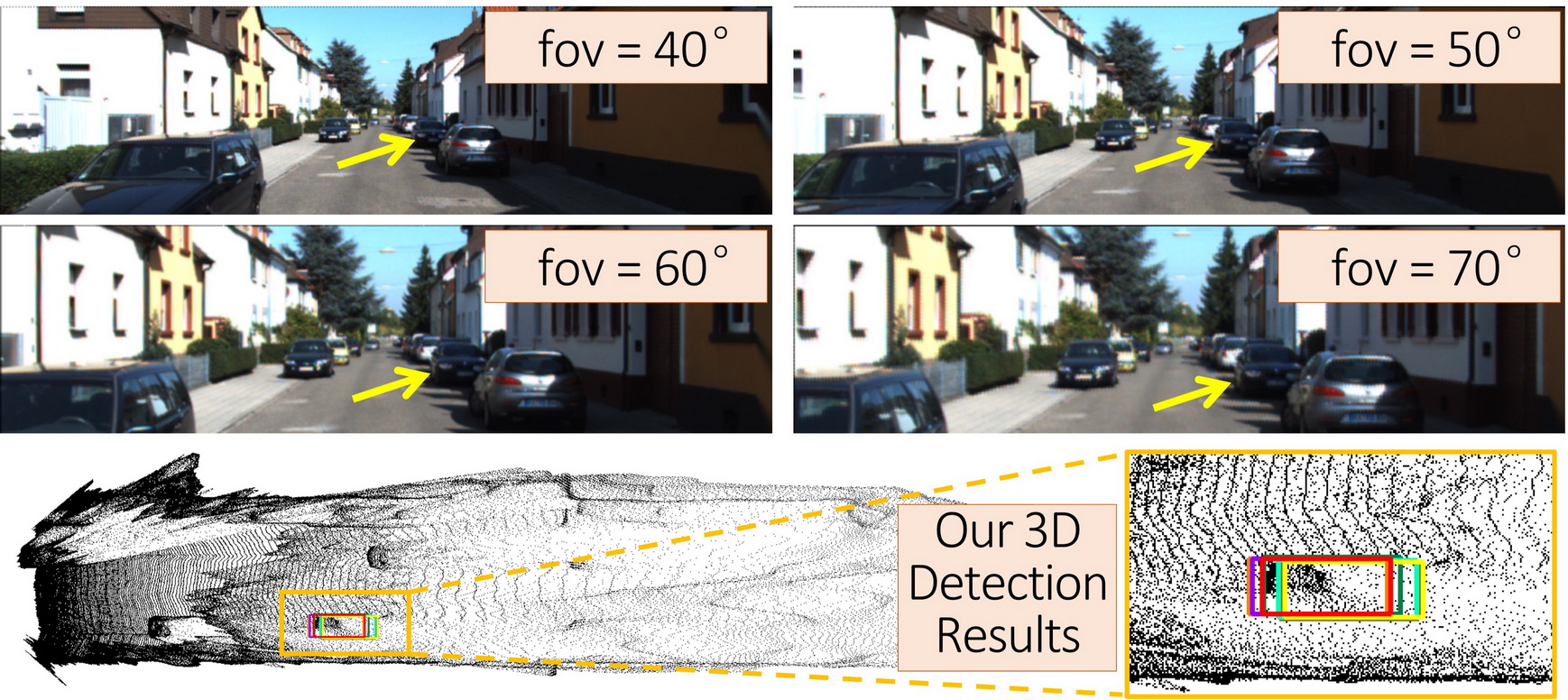}
   \caption{ 3D detection results in bird-eye view with different intrinsic parameters. The red box indicates the ground-truth and the boxes in other colors indicate our results. Our detection results are very closely to the ground-truth in the 3D space though the visual 2D images are different.}
   \label{Fig:ComparisonDetection}
   \end{figure}

   To justify the robustness of our 2D/3D parsing approach, we generate images with different intrinsic parameters of the camera while maintaining the same camera extrinsic parameters and vehicle' poses (Fig.~\ref{Fig:ComparisonDetection}). Many SOTA methods (\eg, \cite{ma2019accurate}, \cite{cai2020monocular}, \cite{ding2020learning}) fail in such cases because they are trained by the depth with a fixed camera parameters. As a result, they predict these vehicles with varying poses, which follow perspective projection (big for near and small for far). In contrast, our method outputs the canonical 3D points which are ono-2-one mapped to image pixels, and then computes the 6-DoF poses using the RANSAC-PnP algorithm. Fig.~\ref{Fig:ComparisonDetection} shows that our 3D detection results are accurate compared to the ground-truth.

   \subsection{Performance Analysis}

   %
   %
   %


   \subsubsection{The Impact of Vehicle Shadow}
   As highlighted in Sec.~\ref{sec:view_syn}, generating a vehicle's shadow is important for improving the quality of synthesized images. Here, we train our network using the synthesized data with and without shadows. Tab.~\ref{tab:shadow} lists the evaluation results. The data synthesized with shadows outperforms the other data on the tasks of 2D detection, instance segmentation, and 6-DoF pose by 1.6\%, 1.5\%, and 2.1\%, respectively.  
   
   \begin{table}[!htbp]
     \begin{center}
     \addtolength{\tabcolsep}{-1.5pt}
      \begin{tabular}{c|c|c}
      \toprule[1pt]
      \textbf{Tasks} & \textbf{w/o Shadow} & \textbf{with Shadow}  \\ 
      \hline
      \textit{2D Detection (mAP}) & 58.3 & \textbf{59.7} \\
      
      \textit{Ins. Seg. (mAP}) & 53.1 & \textbf{54.6}\\	
      
      \textit{6-DoF Pose (Abs-mean}) & 30.9 & \textbf{33.0}\\
      
      \textit{6-DoF Pose (Rel-mean}) & 25.9 & \textbf{27.8}\\
      \bottomrule[1pt]
      \end{tabular}
      \vspace{-0.05in}
       \caption{Ablation study of shadows on 2D/3D tasks. The data synthesized with shadow can effective improve the performance of 2D/3D parsing. }
        \label{tab:shadow}
   
     \end{center}
   \end{table}
   
   \vspace{-0.1in}

   \subsubsection{Diversity of the Synthesized Data}
\begin{figure}
      \centering
      \includegraphics[width=0.98\linewidth]{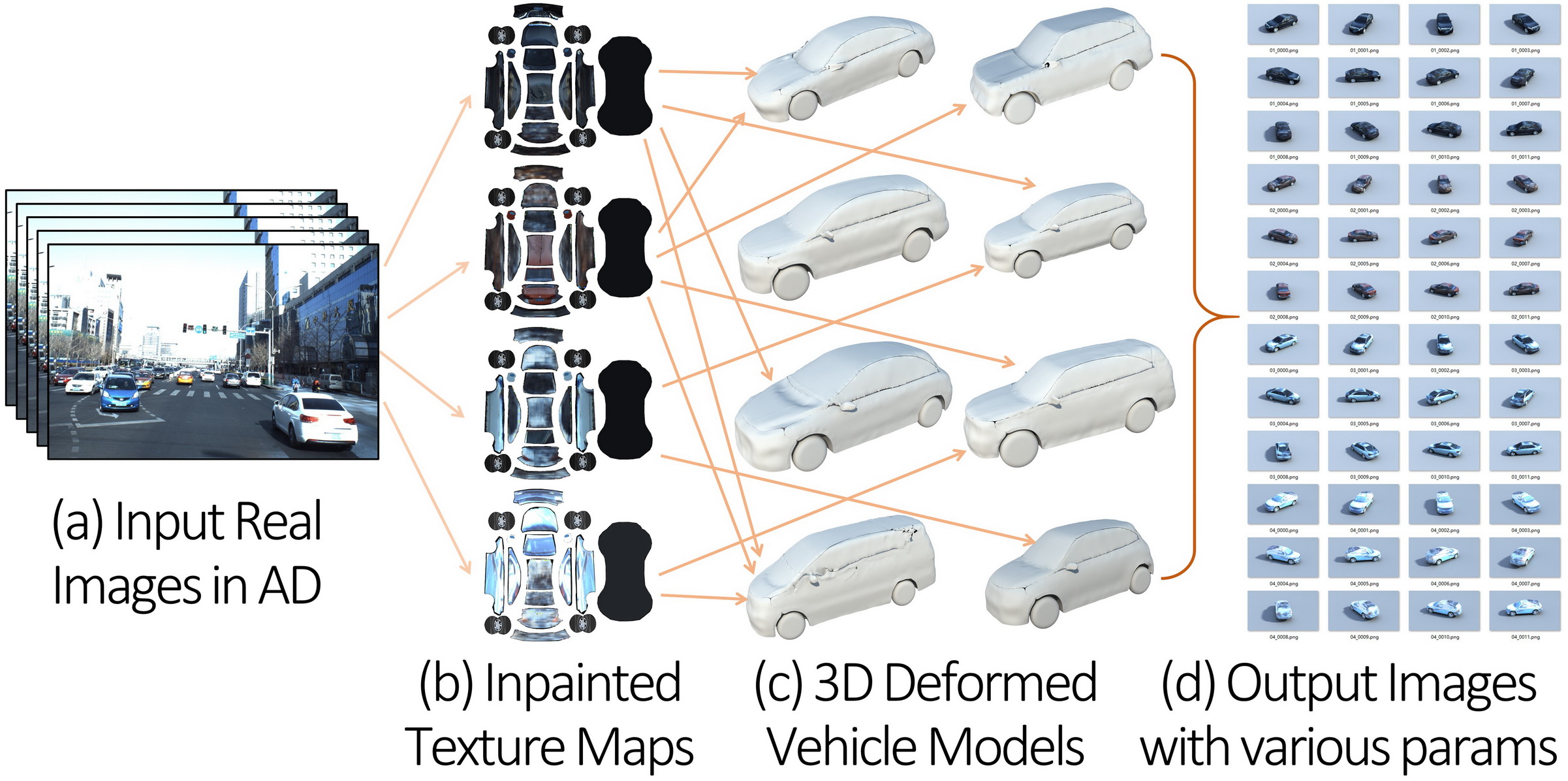}
      \caption{We use the real-traffic images for texture generation and the deformable 3D template for model generation, plus various illuminations for 3D rendering. All of the factors ensure the diversity and fidelity of our synthesized data.}
      \label{Fig:Diversity}
   \end{figure}

   Empirically, the performance of deep network largely relies on the training data (\ie amount and diversity). Our supplement shows the relationship between the data amount and the network performance. Here, we highlight that we can generate diverse data. As shown in Fig.~\ref{Fig:Diversity}, we can synthesize photo-realistic texture maps from the real images. These texture maps are mapped to a deformable vehicle template (\eg, \cite{lu2020permo}) which can be deformed to many vehicle models with different types (\eg, car, SUV, MPV, \etc). Finally, we render these textured models with different intrinsic and extrinsic parameters of cameras under various illuminations.

   \subsubsection{Improvements on the KITTI Dataset}

   In this paper, we synthesize images for vehicle perception in CVIS. To justify the effectiveness and generalizability of our approach, we evaluate on the \textit{KITTI} dataset~\cite{geiger2013vision, qi2019amodal}. Specifically, we mix our synthesized data of CVIS (2593 images) with the existing training data of \textit{KITTI} (7481 images) to train \textit{Mask-RCNN} \cite{he2017mask}. As shown in Tab.~\ref{tab:improve_ad}, the \textit{mAP} values of 2D detection and instance segmentation are improved by 1.4\% and 1.5\%, respectively.

   \begin{table}[!htbp]
      \begin{center}
        \begin{tabular}{c|c|c}
        \toprule[1pt]
        \textbf{Methods}& \textbf{2D Det.}  & \textbf{Ins. Seg.} \\    
        \hline
        \textit{Existing Training Data} & 43.7 & 39.4 \\
        
        \textbf{\textit{Mix Data (Existing + Ours)}} & \textbf{45.1} & \textbf{40.9} \\	
        \bottomrule[1pt]
        \end{tabular}
        \vspace{-0.05in}
        \caption{ Our synthesized data can also improve the 2D detection and instance segmentation performance on \textit{KITTI}. The \textit{mAP} values are the higher the better. }
        \label{tab:improve_ad}
      \end{center}
     \end{table}
   
     \vspace{-0.1in}

   \section{Conclusions}
   
   In this paper, we present a novel approach for 2D/3D vehicle parsing corresponding to novel views for CVIS. Instead of manually annotating data for network training, we proposed a part-assisted view synthesis approach for data augmentation, which can automatically synthesize novel-view images with ground-truth 2D/3D annotations. Moreover, we presented a new approach that combines deep learning and classical 2D/3D registration techniques to perform robust 2D/3D parsing of vehicles, which can be applied to distinct camera parameters. As part of these developing these algorithms for CVIS, we will able to re-utilize the capabilities of prior AD datasets in a novel manner.
   
   
   
   
   
   {\small
   \bibliographystyle{ieee_fullname}
   \bibliography{biobib}
   }
   
   \end{document}